\documentclass[letterpaper,10pt,conference]{ieeeconf}

\IEEEoverridecommandlockouts   

\usepackage{amsmath,amssymb}
\usepackage{graphicx}
\usepackage{booktabs}
\usepackage{multirow}
\usepackage{microtype}
\usepackage{xcolor}
\usepackage{url}
\usepackage{cite}
\usepackage{balance}
\usepackage{algorithm}
\usepackage{algpseudocode}

\usepackage{cuted}
\usepackage{caption}

\usepackage{graphicx}
\usepackage{color}
\usepackage{transparent}
\usepackage{import}
\usepackage{hyperref} 
\usepackage{array}
\usepackage{siunitx}
\usepackage{makecell}

\patchcmd{\abstract}{---\,}{:\ }{}{}
\usepackage{tikz}
\usetikzlibrary{arrows.meta,positioning,calc,fit}

\usepackage{xspace}

\usepackage{adjustbox}
\usepackage{threeparttable}

\newcommand{\method}{EliGSiR\xspace}
\newcommand{\methodposs}{EliGSiR's\xspace}

\title{\LARGE \bf \method: Continual RGB-D Mapping with Gaussian Splatting under Bounded Compute}

\author{Björn Ellensohn, Elmar Rueckert$^{*}$\thanks{*Equal Advising} and Christian Rauch$^{*}$}

\begin{document}
\bstctlcite{BSTcontrol}
\maketitle

\begin{strip}
    \centering
    \includegraphics[
        width=\textwidth
    ]{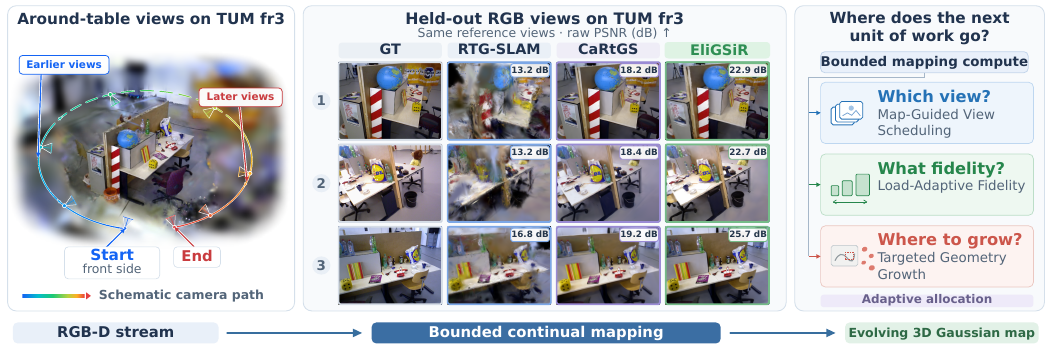}
    \captionof{figure}{\textbf{From a continual RGB-D stream to an evolving 3D Gaussian map.}
    \method allocates bounded compute across view scheduling, supervision fidelity,
    and geometry growth, yielding higher-quality held-out renderings than RTG-SLAM
    and CaRtGS in the shown views.}
    \label{fig:teaser}
    \vspace{-0.5cm}
\end{strip}

\begin{abstract}
Conventional 3D Gaussian Splatting assumes a closed set of observations and long
optimization schedules. Continual RGB-D mapping in contrast poses the problem
that new observations arrive online, while previously reconstructed regions must be preserved.
We present \method{}
\footnote{\href{https://eligsir-project.github.io/}{eligsir-project.github.io}} 
(Evidence-guided Load-adaptive Incremental Gaussian
Splatting with Image Replay), a continual Gaussian mapper that controls how the
available optimization budget is used as the reconstruction evolves.
\textbf{Map-Guided View Scheduling} filters redundant incoming views and
reconsiders retained views according to the current state of the map.
\textbf{Load-Adaptive Fidelity} adjusts supervision resolution to the current
mapping load instead of following a fixed resolution schedule.
\textbf{Targeted Geometry Growth} separates depth supervision from Gaussian
creation and adds geometric capacity only where repeated RGB-D observations
indicate missing or misplaced structure.
Together, these mechanisms adapt which views are optimized, how much image
detail is used, and where the representation grows while mapping remains
active.
We evaluate \method on Replica, TUM RGB-D, ScanNet++, and real RGB-D
sensor sequences, considering both the final reconstruction and the map
available throughout acquisition.
On TUM RGB-D \texttt{fr3/long\_office\_household}, \method reaches
21.52\,dB with the same ground-truth mapping poses used by the controlled
baselines, compared with 19.42\,dB for SplaTAM. In the tracked-pose
comparison, \method with live ORB-SLAM3 poses reaches 23.02\,dB in
155.5\,s, compared with 20.10\,dB in 230.9\,s for CaRtGS using its native
tracker.
We further evaluate reconstruction throughout acquisition and show how \methodposs
adaptive view scheduling, supervision fidelity, and geometry growth improve
the use of the available mapping budget.
\end{abstract}

\section{Introduction}
\label{sec}

Robots operating in real environments require dense 3D representations while
they are moving. Such maps support perception, inspection, navigation, and
manipulation, but they must already be useful before the full environment has
been observed. This differs from offline reconstruction, where all observations
are available in advance and optimization can continue until convergence. A
continual mapper must instead extend its representation as new parts of the
scene become visible while preserving previously reconstructed content.

3D Gaussian Splatting (3DGS)~\cite{kerbl2023gaussians} is especially attractive
for this setting because it combines an explicit scene representation with fast
differentiable rendering. Recent systems have adopted Gaussian
Splatting for RGB-D SLAM and online mapping
~\cite{keetha2024splatam,matsuki2024monogs,peng2024rtgslam,ha2024gsicp}.
These methods demonstrate that Gaussian maps can be built online and, in some cases, in real time. Their optimization must handle a continuous sensor stream introducing new observations.

This creates a resource-allocation problem inside the mapping process.
Compute spent on a redundant view is unavailable for newly observed or
under-optimized regions. Expensive high-resolution updates can reduce the
number of optimization steps that fit into the available time, while aggressive
map growth increases the representation that must subsequently be optimized.
The mapper must therefore balance revisiting useful observations, choosing an
appropriate supervision fidelity, and extending the representation only where
additional geometric support is needed.

We present \method, a continual RGB-D Gaussian mapper designed around these three runtime decisions (Fig.~\ref{fig:teaser}). \method continuously adapts
where its optimization budget is spent as both the reconstruction and the
current workload change. Instead of committing these decisions once when an
observation arrives, the mapper can revisit earlier choices and redirect
computation toward parts of the reconstruction that currently benefit from
further optimization.


Our main contributions are:
\begin{itemize}
\item \textbf{Map-Guided View Scheduling} combines inexpensive
redundancy filtering with continued reprioritization of retained views
according to the evolving reconstruction.

\item \textbf{Load-Adaptive Fidelity} makes supervision resolution
a reversible runtime decision driven by the current mapping load rather
than a predefined resolution schedule.

\item \textbf{Targeted Geometry Growth} decouples depth supervision
from Gaussian creation and introduces additional capacity where repeated
RGB-D observations indicate missing or misplaced geometry.

\end{itemize}

\section{Related Work}
\label{sec:related_work}

\textbf{Online Gaussian SLAM and mapping.}
3D Gaussian Splatting has been adopted by several systems for online
reconstruction and SLAM. SplaTAM~\cite{keetha2024splatam} jointly tracks the
camera and reconstructs an RGB-D scene using silhouette-guided map expansion
and keyframe-based optimization. MonoGS~\cite{matsuki2024monogs} couples
tracking and mapping through differentiable Gaussian rendering, while
Gaussian-SLAM~\cite{yugay2023gaussianslam},
Photo-SLAM~\cite{huang2024photoslam},
GS-SLAM~\cite{yan2024gsslam},
RTG-SLAM~\cite{peng2024rtgslam}, and
GS-ICP-SLAM~\cite{ha2024gsicp} combine Gaussian reconstruction with different
tracking and map-update strategies. LoopSplat~\cite{zhu2025loopsplat}
additionally addresses global consistency through Gaussian submap
registration.
They demonstrate the feasibility of online Gaussian map reconstruction,
but differ in how compute resources are allocated and how views are scheduled
for optimization.

\method focuses specifically on how this
mapping budget can be redistributed as the reconstruction changes.

\textbf{View selection and replay.}
Selecting a limited set of observations for repeated optimization is common in
dense neural mapping. NICE-SLAM~\cite{zhu2022niceslam} and
Point-SLAM~\cite{sandstrom2023pointslam} use keyframes to revisit earlier
observations while incorporating new input. Online Gaussian mappers
retain selected historical views and use recency, motion, overlap, or
covisibility to control which observations participate in optimization.
This avoids redundant work, but neglects that earlier views can become
useful again as the reconstruction evolves.
\textbf{Map-Guided View Scheduling}
therefore keeps retained views available for continued reprioritization instead
of treating their initial admission decision as permanent. Spatial
reconstruction error further allows replay to be directed toward parts of the
map that currently require additional optimization.

\textbf{Compute-aware optimization and supervision.}
Real-time Gaussian mapping has motivated several strategies for reducing or
redistributing optimization cost. RTG-SLAM~\cite{peng2024rtgslam} separates
stable and unstable Gaussian populations and restricts optimization and
rendering to the unstable part of the representation.
CaRtGS~\cite{feng2025cartgs} addresses computational imbalance by adapting
training effort and densification during online Gaussian optimization.
Photo-SLAM~\cite{huang2024photoslam} uses Gaussian-pyramid training to progress
from coarse to finer image levels as optimization proceeds.
\textbf{Load-Adaptive Fidelity} instead treats supervision resolution
as a reversible runtime variable: image fidelity follows the current mapping
load and can dynamically adjust between coarse and fine levels throughout acquisition rather
than following a predefined progression.

\textbf{Geometry growth from RGB-D observations.}
Depth provides direct evidence for initializing and extending Gaussian maps.
SplaTAM~\cite{keetha2024splatam} adds Gaussians according to silhouette and
depth evidence, while RTG-SLAM~\cite{peng2024rtgslam} introduces primitives in
newly observed regions and where color or depth errors remain large.
GS-SLAM~\cite{yan2024gsslam} also adapts Gaussian expansion during mapping, and
GSFusion~\cite{wei2024gsfusion} combines a TSDF representation with
image-space subdivision to obtain compact Gaussian geometry.
3DGS-MCMC~\cite{kheradmand2024mcmc} provides a mechanism for
relocating, replacing, and densifying Gaussian primitives.
\textbf{Targeted Geometry Growth} differs in the decision that triggers map
expansion: depth can first supervise the existing representation without
requiring new primitives, while additional capacity is introduced only where
repeated RGB-D observations support missing or misplaced geometry.

\textbf{Online optimization under bounded compute.}
The related methods reduce online reconstruction cost through different
combinations of keyframe selection, selective updates, multi-resolution
training, controlled densification, and compact representations. These
strategies establish that real-time Gaussian mapping is feasible, but
address different parts of the computational problem. \method treats the
available mapping budget as a runtime constraint shared between selecting
retained views for optimization, the dynamic resolution for supervision, and
the region for optimizing Gaussians.

\section{Problem Formulation}
\label{sec:problem}

We consider continual RGB-D mapping of a static scene with known
camera poses. At time \(i\), the mapper receives an RGB image \(I_i\), a depth
map \(D_i\), camera intrinsics \(K_i\), pose \(T_{wc,i}\), and timestamp
\(\tau_i\), constituting an observation $o_i = (I_i, D_i, K_i, T_{wc,i}, \tau_i)$.
Only arrived observations can update the current Gaussian map \(\Theta_i\).
Camera poses may be provided by a reference trajectory or a
tracking frontend.

The observation stream and map optimization proceed at different rates. Under
bounded compute, the mapper cannot assume that every retained view can be
optimized equally and at full resolution. Newly observed
depth does not necessarily justify immediate growth of the Gaussian
representation. The available optimization budget must therefore be distributed
across the runtime decisions about frame scheduling, image fidelity, and map growth.

Earlier observations may be retained for later optimization, to which we refer as \emph{replay}.
Their usefulness can
change as the reconstruction evolves: a view may become less important once its
region is well reconstructed, or more useful again when later updates reveal
remaining errors or degrade previously reconstructed content.

Our objective is to keep the map responsive to the incoming stream while
directing the available compute toward the work that currently benefits the
reconstruction quality most.

\section{Method}
\label{sec:method}

\begin{figure*}[t]
    \centering
    \includegraphics[width=1\linewidth]{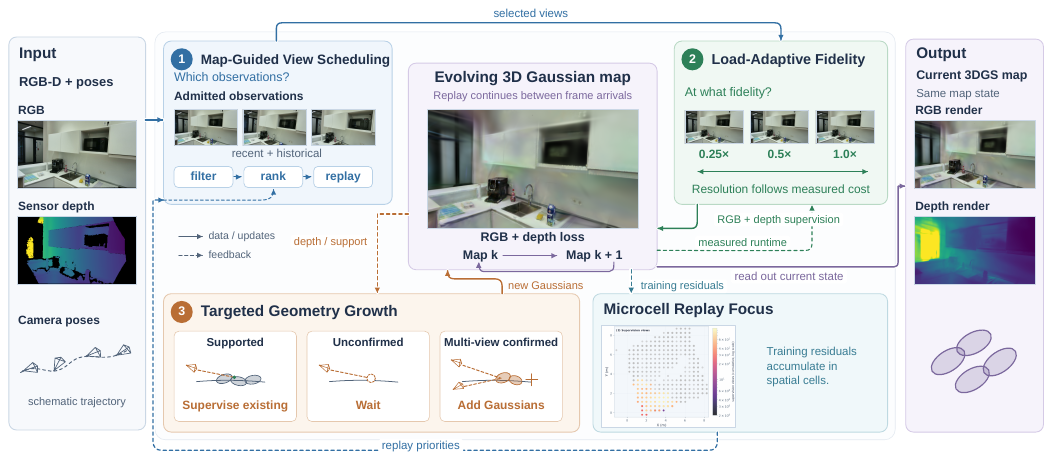}
    \caption{
        \textbf{Method overview.}
        \method allocates mapping work across observation selection and replay,
        supervision fidelity, and geometry growth.
        Retained observations provide recent and historical supervision,
        measured mapping cost controls image resolution, and RGB-D evidence
        guides where the Gaussian representation grows.
    }
    \label{fig:method_overview}
    \vspace{-0.5cm}
\end{figure*}

\method keeps the Gaussian representation available for standard photometric
and depth-based optimization, but changes how the available mapping budget is
allocated while the RGB-D stream is active. It makes three runtime decisions (Figure~\ref{fig:method_overview}):
which views receive optimization, at what image fidelity they are supervised,
and where additional geometric capacity is introduced.
Map-Guided View Scheduling determines which current and retained views are
used for optimization, Load-Adaptive Fidelity controls the cost of their
supervision, and Targeted Geometry Growth adds map capacity where the current
representation is insufficient.

\subsection{Map-Guided View Scheduling}
\label{sec:view_scheduling}

The utility of an observation is not fixed when it enters the map.
An admitted view may be important while its region is first reconstructed,
lose priority once that region is well explained, and become useful again if
later optimization exposes remaining errors.
\method therefore separates the initial admission of a view from its
longer-term priority for supervision.

\textbf{Initial filtering and admission.}
Incoming observations first pass a lightweight admission stage.
A pose-based pre-filter removes redundant views before map-based
evaluation.
The remaining candidates are compared with the current Gaussian map, using
view overlap to identify observations that provide additional coverage.
If no suitable view has been admitted for a fixed interval, the next suitable
candidate is retained through a maximum-gap rule.
This keeps the admission stage inexpensive while avoiding a fixed-stride
keyframe policy.

\textbf{Continuous supervision-set refinement.}
Admission does not assign a permanent keyframe status.
\method maintains an active supervision set together with lower-priority
retained views.
The active set contains observations currently selected for repeated
optimization, while retained views remain available for later replay.

These priorities are updated as the reconstruction changes.
A retained view can return to active supervision when it becomes more useful,
and can lose priority again once the region it covers is
sufficiently reconstructed.
Historical observations therefore remain available for later optimization
instead of being discarded after their initial ranking.

\textbf{Spatial supervision focus.}
View-level scheduling alone can still spend unnecessary compute on already well reconstructed regions.
\method therefore divides the observed scene into fixed world-space
microcells and accumulates reconstruction error from training renders that are
already produced during optimization.
The resulting spatial signal indicates which parts of the map remain weak or
have regressed.

Views covering these regions receive increased replay priority, while views
dominated by well-reconstructed regions are selected less often.
This allows supervision to shift toward parts of the scene that currently need
additional optimization without losing access to the retained observation
history.

\subsection{Load-Adaptive Fidelity}
\label{sec:adaptive_fidelity}

Selecting a useful view does not determine how much compute should be spent on
its optimization.
Full-resolution supervision preserves the most image detail, but also makes
each update more expensive.
When newly observed scene content increases the amount of mapping work,
performing every update at full resolution can reduce the number of
optimization steps that fit into the available time.
\method therefore treats supervision fidelity as a runtime decision.

The scheduler measures the recent mapping cost and selects
from a small set of discrete image resolutions, prioritising high resolutions
when sufficient compute is available, and reduces resolution as mapping load increases.

The decision is reversible: supervision can return to higher resolution when
the mapping load drops and decrease again when the workload rises.
Supervision fidelity therefore follows the current mapping load rather than a
predefined coarse-to-fine schedule.

Changing the supervision resolution does not modify the retained source
observation.
The original RGB-D data remains available, and the camera intrinsics are
scaled consistently with the image grid.
\method can therefore reduce supervision cost temporarily without discarding
the image detail available for later optimization.

\subsection{Targeted Geometry Growth}
\label{sec:geometry_growth}

Depth supervision and Gaussian creation serve different purposes.
A depth observation can constrain Gaussians that already represent a surface
without requiring new primitives.
Newly observed or insufficiently represented regions, in contrast, require
additional geometric support.
\method therefore separates the use of depth for supervision from the decision
to grow the representation.

\textbf{Sparse structural support.}
Newly observed space first receives a bounded set of sparse depth-derived
Gaussians.
Rather than densely converting all depth into map geometry,
these initial primitives establish the broad structure required to begin
reconstruction.
A single incoming observation can therefore extend the map without immediately
introducing a dense Gaussian population.
This initial support is not treated as the final geometry.
Retained RGB-D observations remain available after their first use, allowing
the representation to be refined as further observations become available.

\textbf{Progressive geometry growth.}
When a region is revisited, \method compares the observed surface with the
support already provided by the Gaussian map.
Additional insertion is suppressed where the current representation adequately
explains the observations and directed toward regions that remain unsupported
or geometrically misplaced.
Repeated observations therefore provide additional evidence before finer
geometry is introduced, rather than allowing each unsupported depth
measurement to trigger immediate map growth.

Depth consequently has two roles.
It supervises geometry that already exists and provides evidence for extending
regions that remain insufficiently represented.
The first occurs during ordinary optimization; the second changes the capacity
of the representation and is controlled separately.

\textbf{Population refinement.}
We use the energy-based 3DGS-MCMC strategy~\cite{kheradmand2024mcmc} for
Gaussian relocation, replacement, and densification.
MCMC provides the population-update mechanism rather than constituting a
separate contribution of \method.
Together with sparse initial support and later targeted insertion, it allows
the Gaussian population to refine as additional observations and optimization
become available.

The same Gaussian representation remains available after the input stream
ends.
If additional refinement is performed, optimization can continue at higher
fidelity and further refine the Gaussian population.
The online mechanisms therefore reduce unnecessary work during acquisition
without replacing the map with a separate reduced representation.

\section{Evaluation}
\label{sec:evaluation}

\subsection{Experimental Setup}
\label{sec:experimental_setup}

Following common evaluation settings in recent dense and Gaussian SLAM
work~\cite{keetha2024splatam,peng2024rtgslam}, we evaluate on representative
scenes from Replica, TUM RGB-D, and ScanNet++. Specifically, we use Replica
\texttt{room2}~\cite{straub2019replica}, TUM RGB-D
\texttt{fr3/long\_office\_household} and \texttt{fr1/desk}~\cite{sturm2012tum}, Scannet~\cite{dai2017scannet}
and ScanNet++ \texttt{8b5caf3398}~\cite{yeshwanth2023scannetpp}.
Additional cross-scene evaluation uses Replica \texttt{office0} and the
real-sensor Orbbec sequences \texttt{floor2} and \texttt{kitchen1}.
Qualitative cross-dataset evaluation additionally includes ScanNet.
These scenes cover synthetic reconstruction, real handheld RGB-D acquisition,
and high-quality novel-view evaluation while overlapping with sequences used
by the evaluated baselines.
Frames are processed in temporal order, while admitted observations may remain
available for later replay. Optimization can therefore revisit previously
observed views, but never access observations that have not yet arrived.

The main Replica experiments run on an AMD Radeon R9700 GPU, while
the main TUM experiments use an NVIDIA RTX~4070.
\method is implemented in PyTorch and \texttt{gsplat}~\cite{ye2025gsplat},
with CUDA and ROCm support for NVIDIA and AMD GPUs. The same implementation
also runs on integrated platforms such as AMD Strix Halo and is available as
a ROS~2 package. Camera poses are supplied through a generic interface and can
originate from ground truth, visual odometry, or a SLAM frontend. Baselines run
with their original implementations and native rasterizers. For cross-method
evaluation, exported Gaussian maps are additionally rendered through a common
\texttt{gsplat} path.

\subsection{Evaluation Protocol}
\label{sec:evaluation_protocol}

\textbf{Held-Out Views.}
A fixed subset of frames is excluded from map optimization and used for
evaluation. Within each comparison set, all methods use the same held-out
split. During mapping, these views are rendered every 100 optimization updates.

\textbf{Rendering Consistency.}
All held-out evaluations use identical camera poses, intrinsics, and reference
images. Exported maps are checked with both the native rasterizer, where
available, and a common \texttt{gsplat} path to ensure that renderer differences
do not affect the comparison.

\textbf{Pose Sources.}
Evaluation poses are fixed, while mapping poses follow the method setup.
All controlled comparisons use the same RGB-D stream and ground-truth poses.
CaRtGS is the only exception because it does not support externally supplied
poses and therefore uses its native tracking trajectory. For comparison, we
also run \method with live ORB-SLAM3 poses~\cite{campos2021orbslam3}.

\textbf{Map State.}
We report the evaluated map state because methods differ in how much processing
continues after the input stream. \textsc{ME} denotes \method after the stream
and bounded drain, \textsc{NC} the completed native pipeline of RTG-SLAM or
SplaTAM, and \textsc{NT} the short native post-stream tail of CaRtGS.
$+Nk$\,\textsc{PM} denotes \(N\) thousand additional post-mapping updates;
VarSplat's $+30k$\,\textsc{PM} includes global optimization and bundle
adjustment.

\textbf{Mapping-End Quality.}
Mapping-end quality is measured after the input RGB-D sequence and a short
drain of already scheduled work. Evaluation rendering is excluded from the
reported mapping time. We report reconstruction metrics together with elapsed
mapping time, optimizer updates, supervised views, dropped frames, and Gaussian
count where applicable.

For TUM RGB-D, we additionally optimize an offline reference using all
available frames and ground-truth poses to estimate the reconstruction quality
attainable without the online compute constraint. This result is reported
separately and is not treated as an online baseline.

\textbf{Online Mapping Quality.}
Mapping-end metrics do not capture reconstruction quality during acquisition.
We therefore report held-out PSNR over both elapsed wall-clock time and
cumulative supervised views, separating runtime progress from supervision
efficiency. Temporal partitions further reveal whether parts of the sequence
fall behind during mapping.

We also report supervised pixels, obtained by summing the image resolution of
every supervised view over all optimization updates. Values are given in
megapixels (MPix) and quantify the total amount of image-space supervision.

\textbf{Regional quality.}
Global image metrics can hide local reconstruction failures. We therefore
evaluate fixed \(0.5\,\mathrm{m}\) world-space cells. Each cell \(c\) has a
fixed set \(\Omega_c\) of held-out reference samples defined by camera, pixel,
and reference depth.

Eligibility is independent of the mapper. A cell becomes eligible at
\(t_{\mathrm{vis}}(c)\), the first evaluation checkpoint after its surface has
been observed from at least two distinct source views; held-out frames do not
count. It is measurable when at least 300 reference samples from at least two
held-out views are available. We denote cells that are both eligible and
measurable by
$\mathcal{M}(t)
=
\left\{
c \mid
c \text{ is eligible and measurable at } t
\right\}$.
All regional quality metrics below are computed over \(\mathcal{M}(t)\).

For each measurable cell, we compute RGB mean squared error
\[
e_c(t)=
\frac{1}{3|\Omega_c|}
\sum_{(v,u)\in\Omega_c}
\sum_{k\in\{R,G,B\}}
\left(
\hat I^k_{t,v}(u)-I^{\mathrm{ref},k}_v(u)
\right)^2,
\]
and convert it to cell-level PSNR,
\[
q_c(t)=-10\log_{10}\!\left(\max\{e_c(t),\varepsilon\}\right),
\qquad \varepsilon=10^{-12},
\]
with RGB values normalized to \([0,1]\).

We parameterize acquisition progress by
\[
p(t)=\frac{N_{\mathrm{released}}(t)}{N_{\mathrm{total}}},
\]
where \(N_{\mathrm{released}}(t)\) is the number of source frames released by
checkpoint \(t\).

Causal Visible Quality (CVQ) is the mean cell-level PSNR over the currently
evaluable scene:
\[
\mathrm{CVQ}(t)=
\frac{1}{|\mathcal{M}(t)|}
\sum_{c\in\mathcal{M}(t)} q_c(t).
\]

Causal Usable Coverage (CUC@Q) is the fraction of evaluable cells above a
quality threshold \(Q\):
\[
\mathrm{CUC}@Q(t)=
\frac{|\{c\in\mathcal{M}(t)\mid q_c(t)\ge Q\}|}
{|\mathcal{M}(t)|}.
\]
We use \(Q=20\,\mathrm{dB}\).

To summarize reconstruction quality throughout acquisition, we integrate CVQ
over normalized source progress:
\[
\mathrm{CVQ\text{-}AUC}=
\frac{\int_{p_{\min}}^{p_{\max}}\mathrm{CVQ}(p)\,dp}
{p_{\max}-p_{\min}}.
\]
The cell partition, visibility criteria, and quality threshold are fixed for
all compared methods; missing renders count as failures.

\section{Results}
\label{sec:results}

We first compare final-map reconstruction across benchmark and real-sensor
sequences. We then evaluate reconstruction quality during acquisition and
continued refinement, isolate the effects of the three runtime decisions, and
finally provide qualitative comparisons.

The final-map comparison, causal ablations, and refinement study use separate
runs and evaluation splits; absolute PSNR values should therefore not be
compared across these experiment blocks.

\subsection{Final-Map Reconstruction}
\label{sec:results_controlled}

\begin{table}[t]
\centering
\begin{threeparttable}

\caption{
\textbf{Reconstruction quality and processing time across RGB-D scenes.}
Methods are reported at their evaluated map states using their native
finalization procedures. Final-map quality and elapsed mapping time should be
considered jointly, since some native pipelines continue substantial
optimization after acquisition. Reported times are descriptive because hardware
and timing scopes differ across experiments.
}
\label{tab:final_reconstruction}

\setlength{\tabcolsep}{6pt}
\begin{tabular}{llcccc}
\toprule
Method &
Map state &
PSNR$\uparrow$ &
SSIM$\uparrow$ &
LPIPS$\downarrow$ &
Time [s] \\
\midrule

\multicolumn{6}{l}{\emph{TUM RGB-D \texttt{fr3/long\_office\_household}: GT mapping poses}} \\
SplaTAM  & \textsc{NC}             & 19.42 & 0.703 & 0.445 & 1382.6 \\
RTG-SLAM & \textsc{NC}             & 14.46 & 0.553 & 0.614 & 1147.8 \\
VarSplat & $+30k$\,\textsc{PM}     & 16.44 & 0.574 & 0.510 & 2495.5 \\
\method  & \textsc{ME}             & \textbf{21.52} & \textbf{0.757} & \textbf{0.320} & \textbf{176.1} \\
\addlinespace[1pt]

\multicolumn{6}{l}{\emph{TUM RGB-D \texttt{fr3/long\_office\_household}: tracked poses}} \\
CaRtGS                  & \textsc{NT} & 20.10 & 0.689 & 0.363 & 230.9 \\
\method{}   & \textsc{ME} & \textbf{23.02} & \textbf{0.803} & \textbf{0.276} & \textbf{155.5} \\
\midrule

\multicolumn{6}{l}{\emph{Replica \texttt{office0}}} \\
RTG-SLAM & \textsc{NC}         & 34.62 & 0.940 & 0.263 & 412.4 \\
VarSplat & $+30k$\,\textsc{PM} & \textbf{40.91} & \textbf{0.972} & \textbf{0.203} & 3742.2 \\
\method  & \textsc{ME}           & 34.20 & 0.938 & 0.223 &
\textbf{324.0}\tnote{a} \\
\addlinespace[1pt]

\multicolumn{6}{l}{\emph{Replica \texttt{room2}}} \\
\method & $+2k$\,\textsc{PM} & 37.07 & 0.964 & 0.137 & 746.4 \\
\addlinespace[1pt]

\multicolumn{6}{l}{\emph{Orbbec \texttt{floor2}}} \\
RTG-SLAM & \textsc{NC}         & 18.52 & 0.718 & 0.461 & 775.2 \\
SplaTAM  & \textsc{NC}         & 21.40 & 0.782 & 0.390 & 18349.6 \\
VarSplat & $+30k$\,\textsc{PM} & \textbf{25.88} & \textbf{0.883} & \textbf{0.254} & 2274.7 \\
\method  & \textsc{ME}         & 16.27 & 0.601 & 0.510 & \textbf{555.1} \\
\addlinespace[1pt]

\multicolumn{6}{l}{\emph{Orbbec \texttt{kitchen1}}} \\
RTG-SLAM & \textsc{NC} & 16.75 & 0.669 & 0.506 & 408.0 \\
SplaTAM  & \textsc{NC} & 17.59 & 0.731 & 0.447 & 2456.7 \\
CaRtGS   & \textsc{NT} & 20.12 & \textbf{0.856} & 0.281 & \textbf{99.2}\tnote{b} \\
\method  & \textsc{ME} & \textbf{23.50} & 0.840 & \textbf{0.248} & 203.3 \\
\addlinespace[1pt]

\multicolumn{6}{l}{\emph{ScanNet++ \texttt{8b5caf3398}}} \\
\method & \textsc{ME} & 19.58 & 0.817 & 0.307 & 390.9 \\
\addlinespace[1pt]

\multicolumn{6}{l}{\emph{TUM RGB-D \texttt{fr1/desk}}} \\
RTG-SLAM & \textsc{NC} & 16.07 & 0.559 & 0.570 & 707.0 \\
SplaTAM  & \textsc{NC} & 18.75 & 0.659 & 0.471 &
4726.5 \\
\method  & $+30k$\,\textsc{PM}       & \textbf{20.60} & \textbf{0.729} & \textbf{0.408} & \textbf{678.3} \\

\bottomrule
\end{tabular}

\begin{tablenotes}[flushleft]
\footnotesize
\item[a] Timestamp-derived. The Replica \texttt{office0} time additionally
includes evaluation/video generation.
\item[b] Mapping time only; excludes the per-view pose refinement used to
obtain the reported CaRtGS reconstruction score.
\end{tablenotes}

\end{threeparttable}
\vspace{-0.5cm}
\end{table}

Table~\ref{tab:final_reconstruction} summarizes final-map reconstruction on
the controlled TUM RGB-D comparison and additional benchmark and real-sensor
sequences.

On TUM RGB-D \texttt{fr3/long\_office\_household}, the controlled comparison
separates common ground-truth mapping poses from tracked poses. With the same
ground-truth poses, \method reaches 21.52\,dB, compared with 19.42\,dB for
SplaTAM, 16.44\,dB for VarSplat, and 14.46\,dB for RTG-SLAM. With live
tracking, \method reaches 23.02\,dB, compared with 20.10\,dB for CaRtGS using
its native tracker.

\begin{figure}[t]
    \centering
    \includegraphics[
        width=\linewidth,
        trim=0px 0px 0px .9cm,clip
    ]{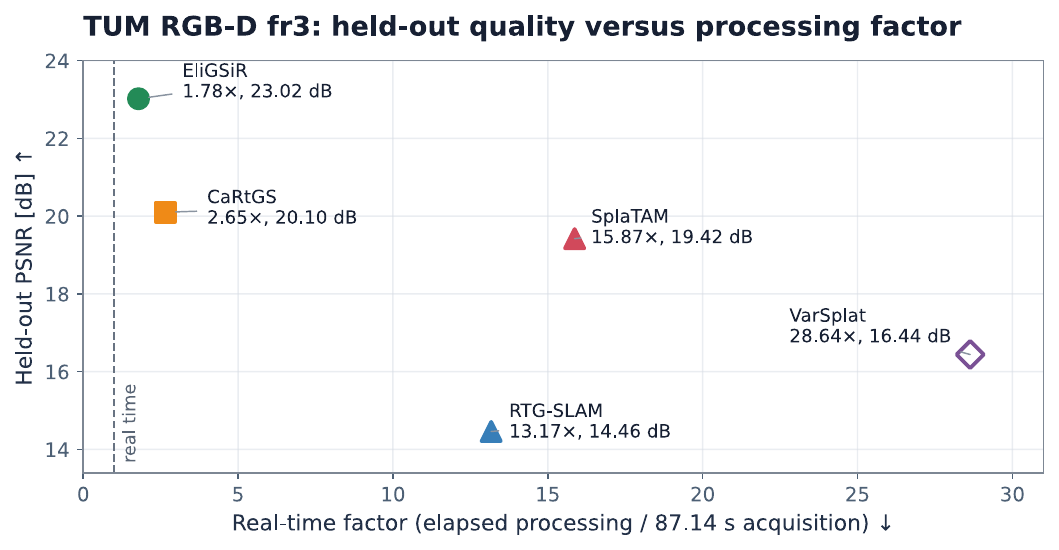}
    \caption{
        \textbf{Held-out reconstruction quality versus real-time factor on
        TUM RGB-D \texttt{fr3/long\_office\_household}.}
        RTF is elapsed processing time divided by the 87.14\,s acquisition
        duration; $1\times$ denotes real-time processing.
        SplaTAM, RTG-SLAM, and VarSplat use the common ground-truth mapping
        poses, while CaRtGS uses its native tracker and \method uses live
        ORB-SLAM3 poses.
    }
    \label{fig:fr3_rtf}
    \vspace{-0.3cm}
\end{figure}

Figure~\ref{fig:fr3_rtf} shows the trade-off between reconstruction quality
and processing time. With live tracking, \method reaches 23.02\,dB at a
real-time factor of 1.78$\times$, compared with 20.10\,dB at 2.65$\times$
for CaRtGS. The GT-pose baselines require 13.17--28.64$\times$ the sequence
duration to reach their reported final maps. Their final PSNR should therefore
be read together with the processing time needed to obtain it.

Across the additional scenes, \method reaches 23.50\,dB on
\texttt{kitchen1}, 37.07\,dB on Replica \texttt{room2}, and 19.58\,dB on
ScanNet++ \texttt{8b5caf3398}.

\subsection{Quality During Mapping and Continued Refinement}
\label{sec:results_online_quality}

\begin{figure}[t]
    \centering
    
    \includegraphics[
        width=\linewidth,
        trim=13pt 7pt 0pt 0pt,
        clip
    ]{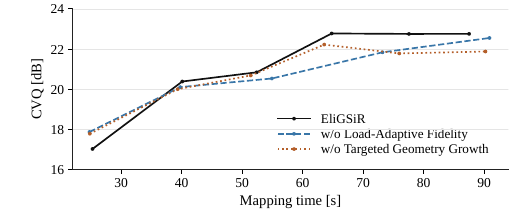}

    \vspace{1mm}

    \caption{
    \textbf{Reconstruction quality during mapping on TUM RGB-D
    \texttt{fr3/long\_office\_household}.}
    CVQ over mapping time for \method,
    w/o Load-Adaptive Fidelity, and w/o Targeted Geometry Growth.
    The corresponding source-progress CVQ-AUC values are 21.18, 20.62, and 20.84\,dB, respectively.
    }
    \label{fig:quality_during_mapping}
    \vspace{-0.7cm}
\end{figure}

We next evaluate reconstruction quality throughout acquisition.
Figure~\ref{fig:quality_during_mapping} compares the full method and its
ablations over mapping time.

The full method reaches a CVQ-AUC of 21.18\,dB, compared with
20.62\,dB without Load-Adaptive Fidelity and 20.84\,dB without
Targeted Geometry Growth.
Although fixed native-resolution supervision reaches a higher final global
PSNR, the full method achieves a higher CVQ-AUC while using substantially less
image-space supervision. In this run, it also completes mapping in 89.15\,s
rather than 103.34\,s.
The workload and fidelity adaptation responsible for this reduction are
shown in Fig.~\ref{fig:load_adaptive_fidelity}.

\begin{table}
    \centering
    \sisetup{detect-weight=true, mode=text}
    \newcolumntype{C}[1]{>{\centering\arraybackslash}p{#1}}
    \newcommand{\head}[1]{\multicolumn{1}{C{0.75cm}}{#1}}  
    \newcommand{\best}[1]{\bfseries #1}
    \setlength{\tabcolsep}{3pt}
    \caption{
        \textbf{Extended optimization on TUM RGB-D
        \texttt{fr3/long\_office\_household}.}
        Separate 4k, 8k, and 16k refinement runs use a 45-view split distinct from Table~\ref{tab:final_reconstruction}. RGB metrics use the original holdout poses after training-pose alignment. Depth RMSE uses representative retained views.
    }
    \begin{tabular}{
        l
        S[table-format=5.0]
        S[table-format=2.2]
        S[table-format=1.3]
        S[table-format=1.3]
        S[table-format=1.3]
        S[table-format=3.1]
        S[table-format=4.1]
    }
        \toprule
        Map state & \head{Updates} & \head{PSNR$\uparrow$} & \head{SSIM$\uparrow$} &
        \head{LPIPS$\downarrow$} & \head{Depth RMSE [m]$\downarrow$} &
        \head{Gaus. [k]} & \head{Total time [s]} \\
        \midrule
        \textsc{ME} &  1372 & 23.27        & 0.807        & 0.269        & 0.459        & 328.1 &  144.9 \\
        +4k\,\textsc{PM}    &  5372 & 23.64        & 0.814        & 0.264        & 0.356        & 356.6 & 1166.5 \\
        +8k\,\textsc{PM}    &  9351 & 24.44        & 0.837        & 0.235        & \best{0.347} & 367.5 & 2188.6 \\
        +16k\,\textsc{PM}   & 17102 & \best{25.62} & \best{0.855} & \best{0.213} & 0.352        & 381.3 & 5062.1 \\
        \bottomrule
    \end{tabular}
    \label{tab:refinement}
    \vspace{-0.1cm}
\end{table}

The same map remains available for continued optimization after acquisition.
Table~\ref{tab:refinement} reports separate refinement runs with 4k, 8k, and
16k additional updates. Appearance quality increases from 23.27\,dB at the
mapping endpoint to 25.62\,dB after 16k updates. Representative-view depth
RMSE improves mainly within the first 4--8k updates and changes little
thereafter.

\subsection{Ablation Study}
\label{sec:results_ablations}

\begin{table}[t]
    \centering
    \caption{
        \textbf{Online ablations across three RGB-D scenes.}
        Endpoint PSNR measures final global quality; CVQ-AUC and CUC@20 measure
        regional quality during acquisition. Supervised pixels and Gaussian count
        report optimization work and map size. Runs are separate from Table~\ref{tab:final_reconstruction}.
    }
    \label{tab:ablations}

    \setlength{\tabcolsep}{6pt}

    \begin{adjustbox}{max width=\columnwidth}
    \begin{tabular}{
        l
        S[table-format=2.2]
        S[table-format=2.2]
        S[table-format=2.1]
        S[table-format=5.1]
        S[table-format=4.1]
    }
        \toprule
        Variant
            & {PSNR$\uparrow$}
            & {\shortstack{CVQ-\\AUC$\uparrow$}}
            & {\shortstack{CUC\\@20$\uparrow$}}
            & {\shortstack{Sup. px.\\{[M]}$\downarrow$}}
            & {\shortstack{Gaus.\\{[k]}}} \\
        \midrule

        \multicolumn{6}{@{}l}{
            \textbf{TUM RGB-D \texttt{fr3/long\_office\_household}}
        } \\[-1pt]

        \method
            & 18.50
            & \bfseries 21.18
            & \bfseries 69.2
            & 398.2
            & 140.1 \\

        w/o adaptive fidelity
            & \bfseries 19.36
            & 20.62
            & 65.4
            & 2786.9
            & 549.5 \\

        w/o targeted growth
            & 17.78
            & 20.84
            & 63.1
            & \bfseries 330.9
            & 120.5 \\

        \midrule

        \multicolumn{6}{@{}l}{
            \textbf{ScanNet++ \texttt{8b5caf3398}}
        } \\[-1pt]

        \method
            & \bfseries 18.71
            & 16.28
            & \bfseries 23.2
            & 17559.6
            & 366.5 \\

        w/o adaptive fidelity
            & 18.58
            & 16.36
            & 16.4
            & 25613.1
            & 341.5 \\

        w/o targeted growth
            & 18.59
            & \bfseries 16.59
            & 17.7
            & \bfseries 14242.3
            & 1108.8 \\

        \midrule

        \multicolumn{6}{@{}l}{
            \textbf{Orbbec \texttt{kitchen1}}
        } \\[-1pt]

        \method
            & 18.25
            & 19.75
            & 55.3
            & 285.6
            & 123.9 \\

        w/o adaptive fidelity
            & \bfseries 22.91
            & \bfseries 20.44
            & \bfseries 77.9
            & 2267.1
            & 565.0 \\

        w/o targeted growth
            & 18.63
            & 19.55
            & 53.3
            & \bfseries 276.9
            & 70.7 \\

        \bottomrule
    \end{tabular}
    \end{adjustbox}
    \vspace{-0.5cm}
\end{table}

Table~\ref{tab:ablations} isolates Load-Adaptive Fidelity and Targeted
Geometry Growth across three RGB-D scenes. Endpoint PSNR measures final global
quality, while CVQ-AUC and CUC@20 capture regional quality during acquisition.
Supervised pixels and Gaussian count measure optimization work and map size.
The results should therefore be read as quality--compute trade-offs rather
than as a single endpoint ranking.

\begin{figure}[t]
    \centering
    \includegraphics[
        width=1\linewidth,
        trim=0pt 0pt 0pt 0pt,
        clip
    ]{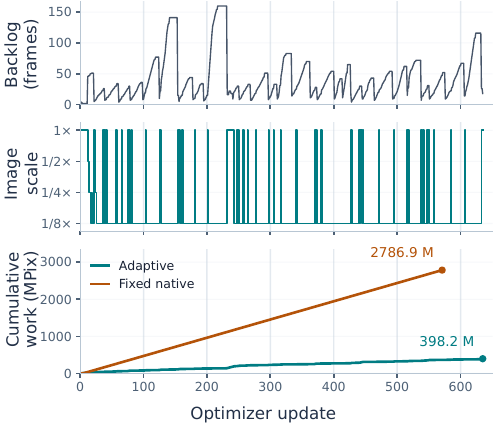}
    \caption{
    \textbf{Load-Adaptive Fidelity on TUM RGB-D
    \texttt{fr3/long\_office\_household}.}
    Mapping backlog, selected supervision scale, and cumulative
    image-space supervision over optimization.
    The controller repeatedly moves between \(1/8\times\),
    \(1/4\times\), \(1/2\times\), and \(1\times\) resolution as
    mapping load changes, reducing cumulative supervision from
    2786.9\,M to 398.2\,M pixels compared with fixed native resolution.
    }
    \label{fig:load_adaptive_fidelity}
    \vspace{-0.5cm}
\end{figure}

\paragraph{Load-Adaptive Fidelity}
Figure~\ref{fig:load_adaptive_fidelity} shows how supervision fidelity follows
the current mapping load. On TUM, adaptive fidelity reduces supervision from
2786.9\,M to 398.2\,M pixels (\(7.0\times\)) and mapping time from
103.34\,s to 89.15\,s, while increasing CVQ-AUC from 20.62 to 21.18\,dB
and CUC@20 from 65.4\% to 69.2\%.
Fixed native resolution reaches the higher endpoint PSNR
(19.36\,dB versus 18.50\,dB), showing that adaptive fidelity improves online
quality for substantially less image-space work rather than maximizing the
final score. Supervised pixels also decrease by \(1.46\times\) on ScanNet++
and \(7.9\times\) on \texttt{kitchen1}.

\paragraph{Targeted Geometry Growth}
Targeted Geometry Growth adds capacity where repeated RGB-D observations show
that the current geometry is insufficient. It increases CUC@20 by 6.1, 5.5,
and 2.0 percentage points on TUM, ScanNet++, and \texttt{kitchen1},
respectively. On ScanNet++, it also reduces the representation from 1.109\,M
to 366.5\,k Gaussians while maintaining similar endpoint PSNR
(18.71\,dB versus 18.59\,dB).

\begin{figure}[t]
    \centering
    \includegraphics[
        width=\linewidth,
        trim=8pt 8pt 8pt 8pt,
        clip
    ]{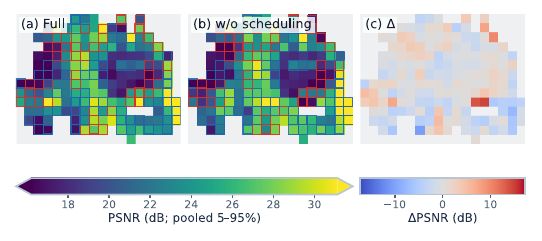}
    \caption{
        \textbf{Spatial effect of Map-Guided View Scheduling.}
        Top-down view of the fixed world-space microcell grid aligned with the
        Gaussian map. Each shown cell summarizes endpoint reconstruction quality for
        the local mapped surface. We compare the full method, the scheduling ablation,
        and their difference
        (\(\Delta\mathrm{PSNR}=\mathrm{Full}-\mathrm{w/o\ scheduling}\))
        over the same observable cells.
        Scheduling leaves aggregate quality similar, but redistributes reconstruction
        quality across the mapped scene.
    }
    \label{fig:scheduling_spatial}
    \vspace{-0.5cm}
\end{figure}

\paragraph{Map-Guided View Scheduling}
Map-Guided View Scheduling redirects optimization toward weak regions.
Figure~\ref{fig:scheduling_spatial} shows the fixed world-space microcell grid
aligned with the Gaussian map, where each cell summarizes quality for the
local mapped surface. Mean and median regional quality remain similar, while
several weak regions improve by more than 10\,dB. The scheduler therefore
redistributes optimization toward regions that still need it rather than
uniformly increasing endpoint quality.

\subsection{Qualitative Results}
\label{sec:results_qualitative}

\begin{figure}[t]
    \centering
    \setlength{\tabcolsep}{1pt}
    \renewcommand{\arraystretch}{0}

    {\scriptsize\textbf{(a) ScanNet}}\par
    \vspace{0.4mm}

    \begin{tabular}{@{}c@{\hspace{0.5mm}}ccc@{}}
        &
        {\tiny View 48} &
        {\tiny View 2624} &
        {\tiny View 5872}
        \\[0.4mm]

        \rotatebox{90}{\tiny Reference} &
        \includegraphics[width=.295\columnwidth]{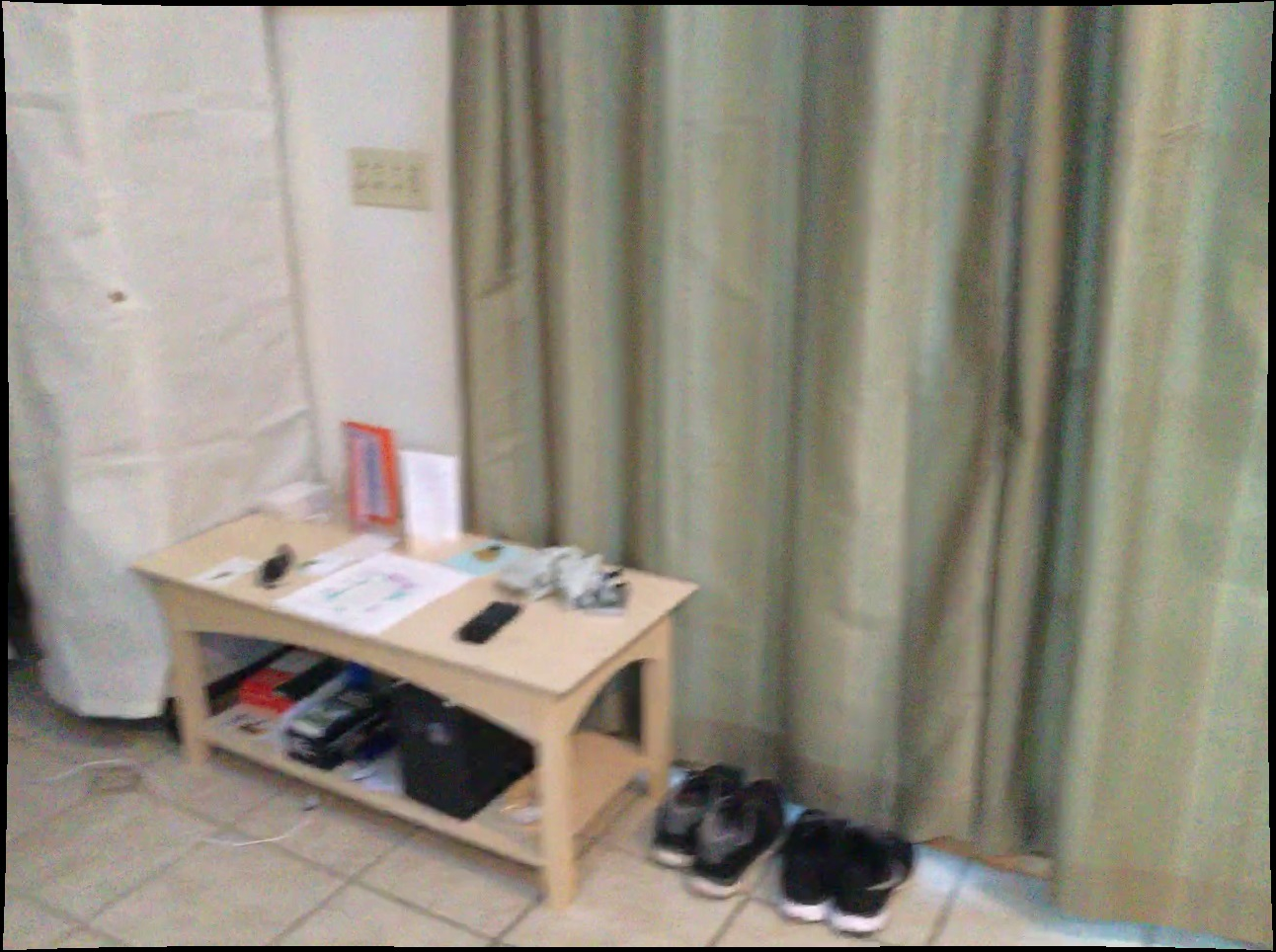} &
        \includegraphics[width=.295\columnwidth]{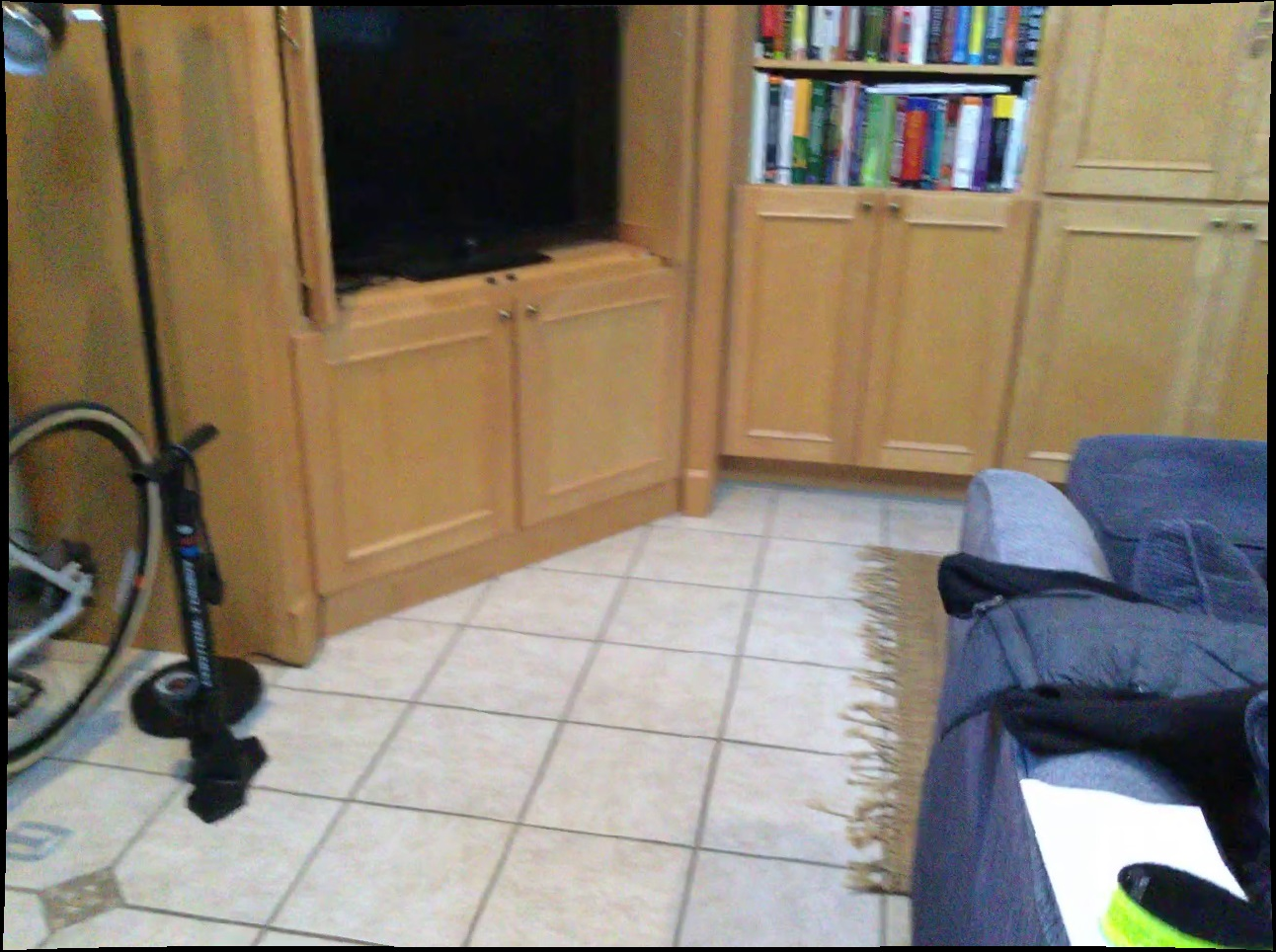} &
        \includegraphics[width=.295\columnwidth]{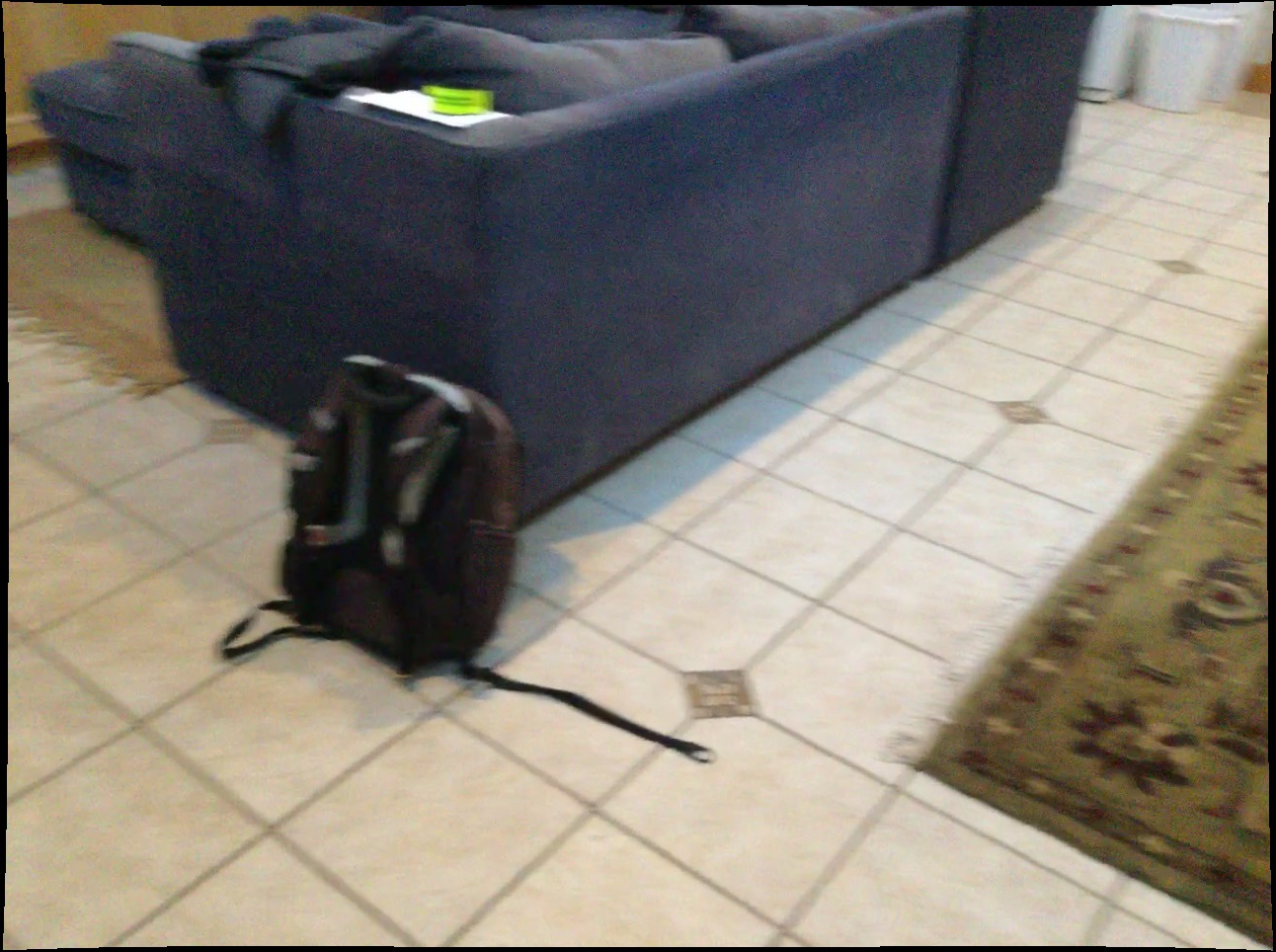}
        \\[-0.5mm]

        \rotatebox{90}{\tiny \method} &
        \includegraphics[width=.295\columnwidth]{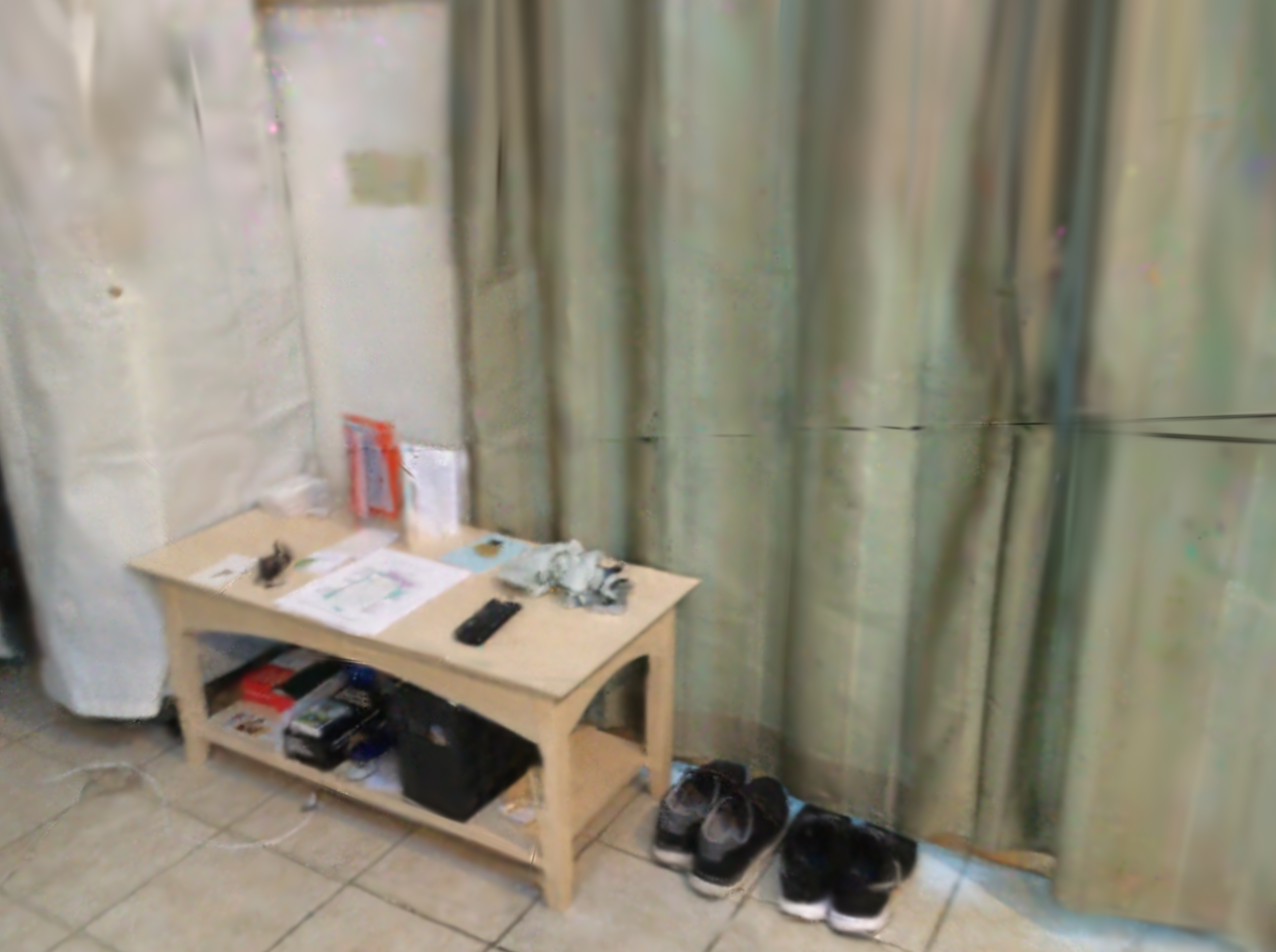} &
        \includegraphics[width=.295\columnwidth]{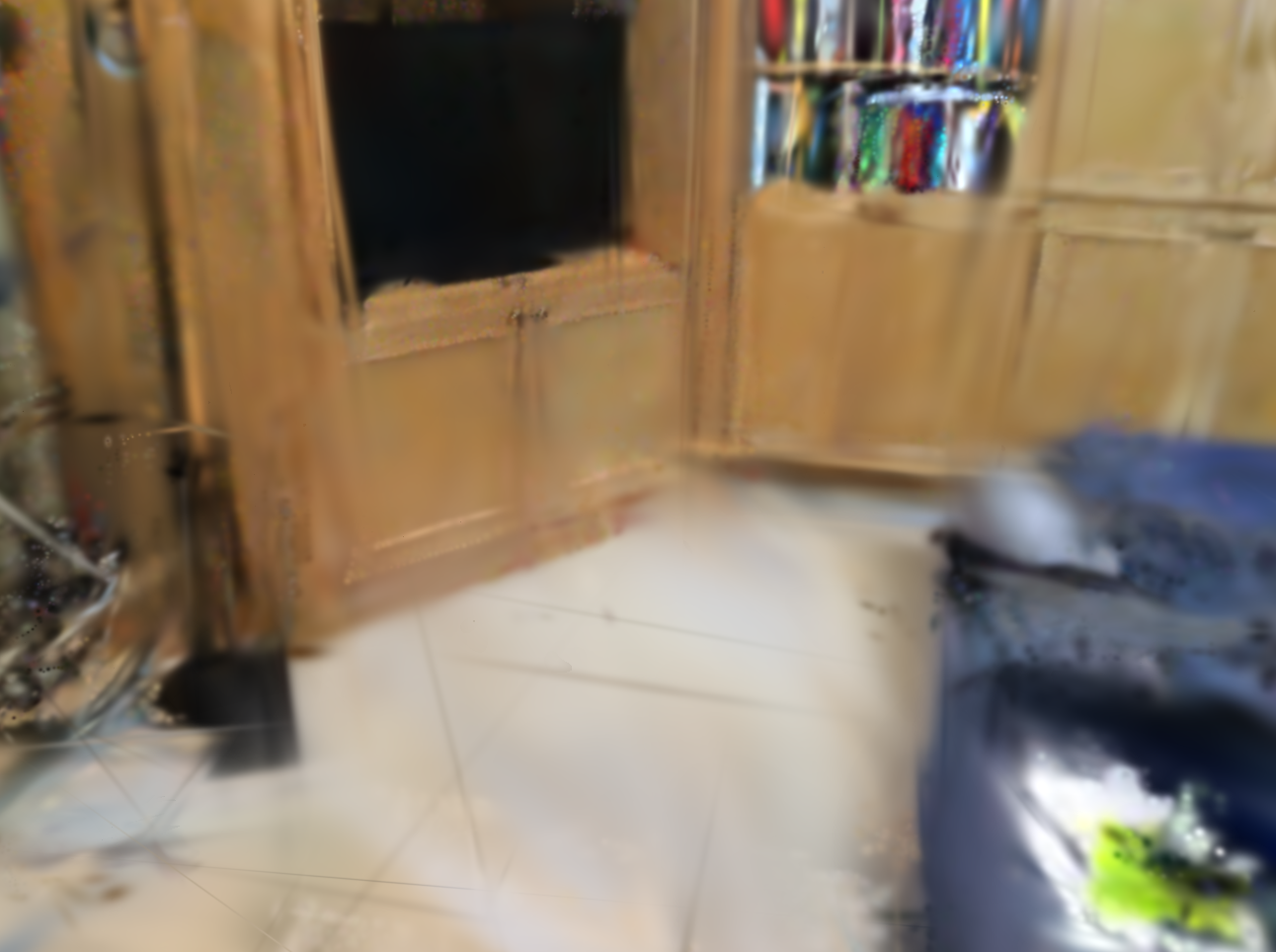} &
        \includegraphics[width=.295\columnwidth]{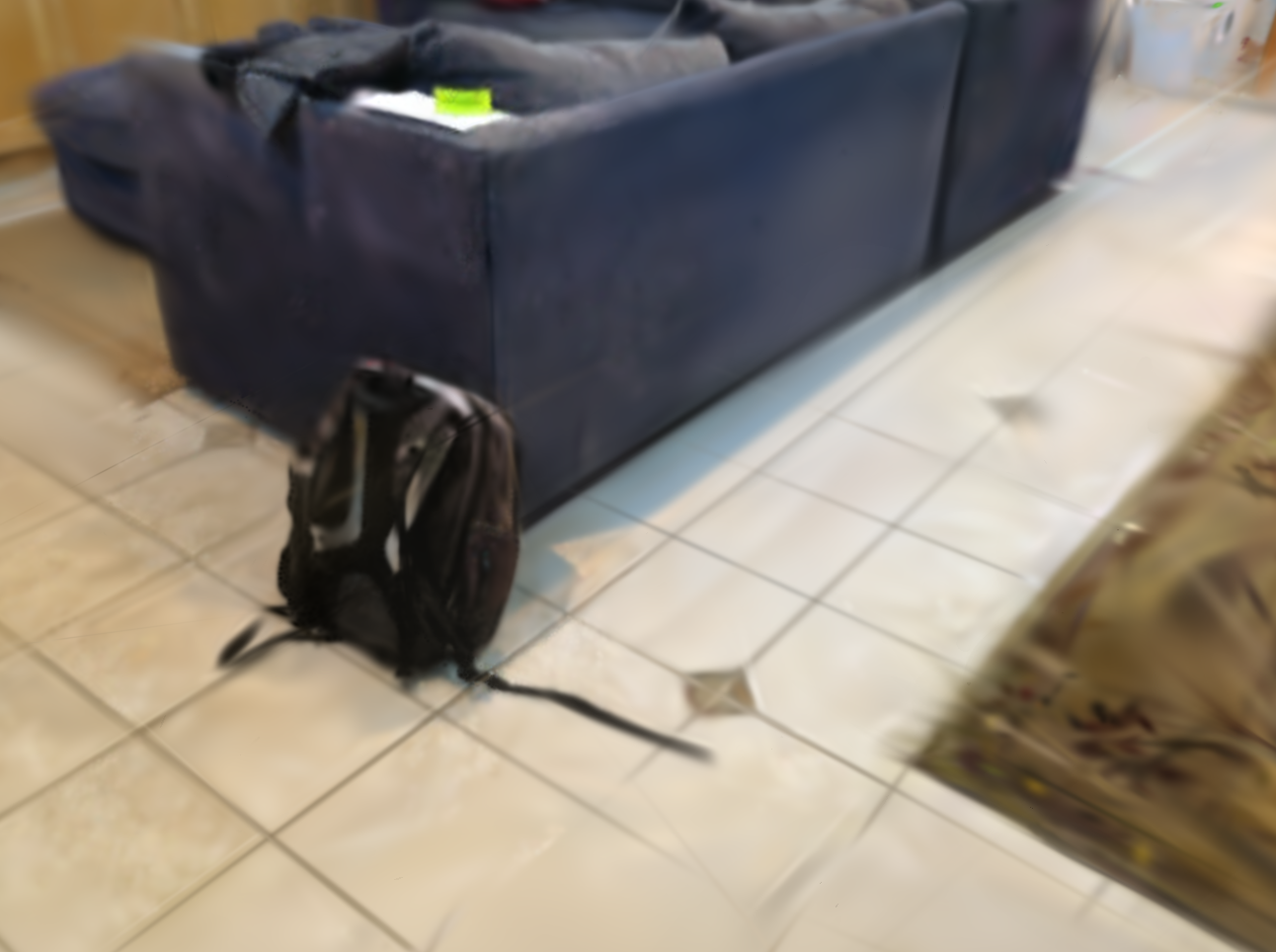}
    \end{tabular}

    \vspace{1.2mm}

    {\scriptsize\textbf{(b) ScanNet++ \texttt{8b5caf3398}}}\par
    \vspace{0.4mm}

    \begin{tabular}{@{}c@{\hspace{0.5mm}}ccc@{}}
        &
        {\tiny View 8} &
        {\tiny View 5624} &
        {\tiny View 6752}
        \\[0.4mm]

        \rotatebox{90}{\tiny Reference} &
        \includegraphics[width=.295\columnwidth]{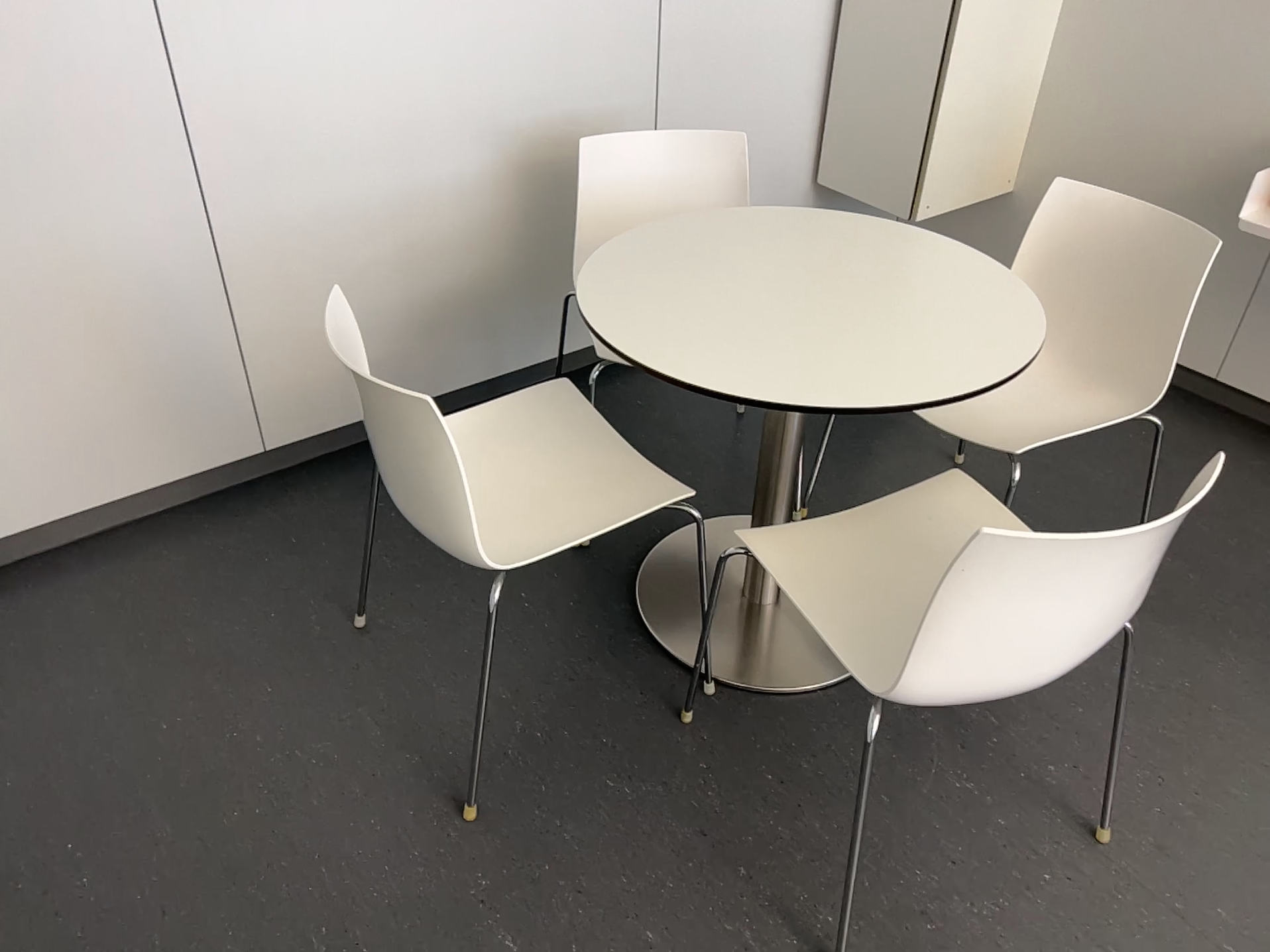} &
        \includegraphics[width=.295\columnwidth]{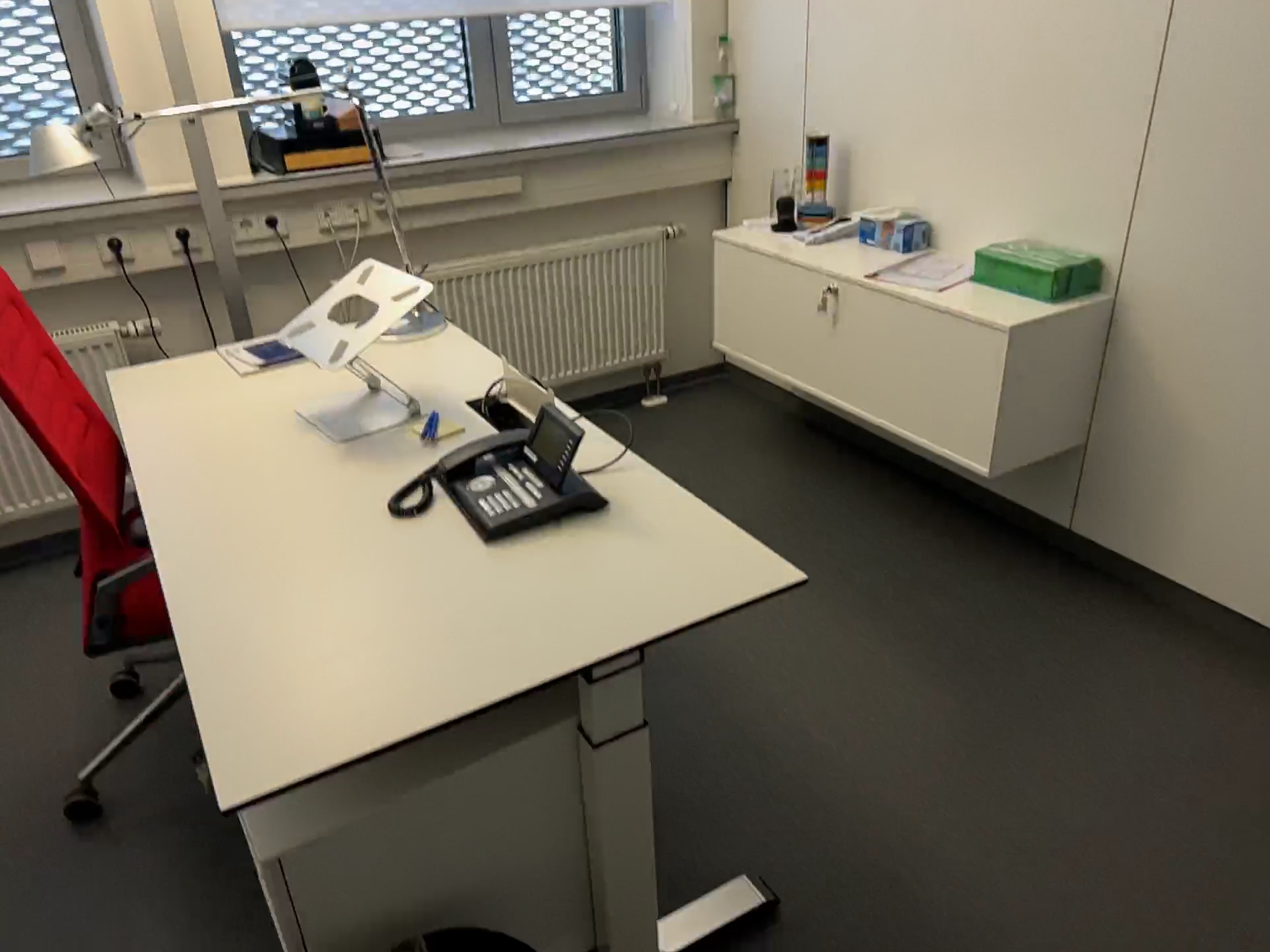} &
        \includegraphics[width=.295\columnwidth]{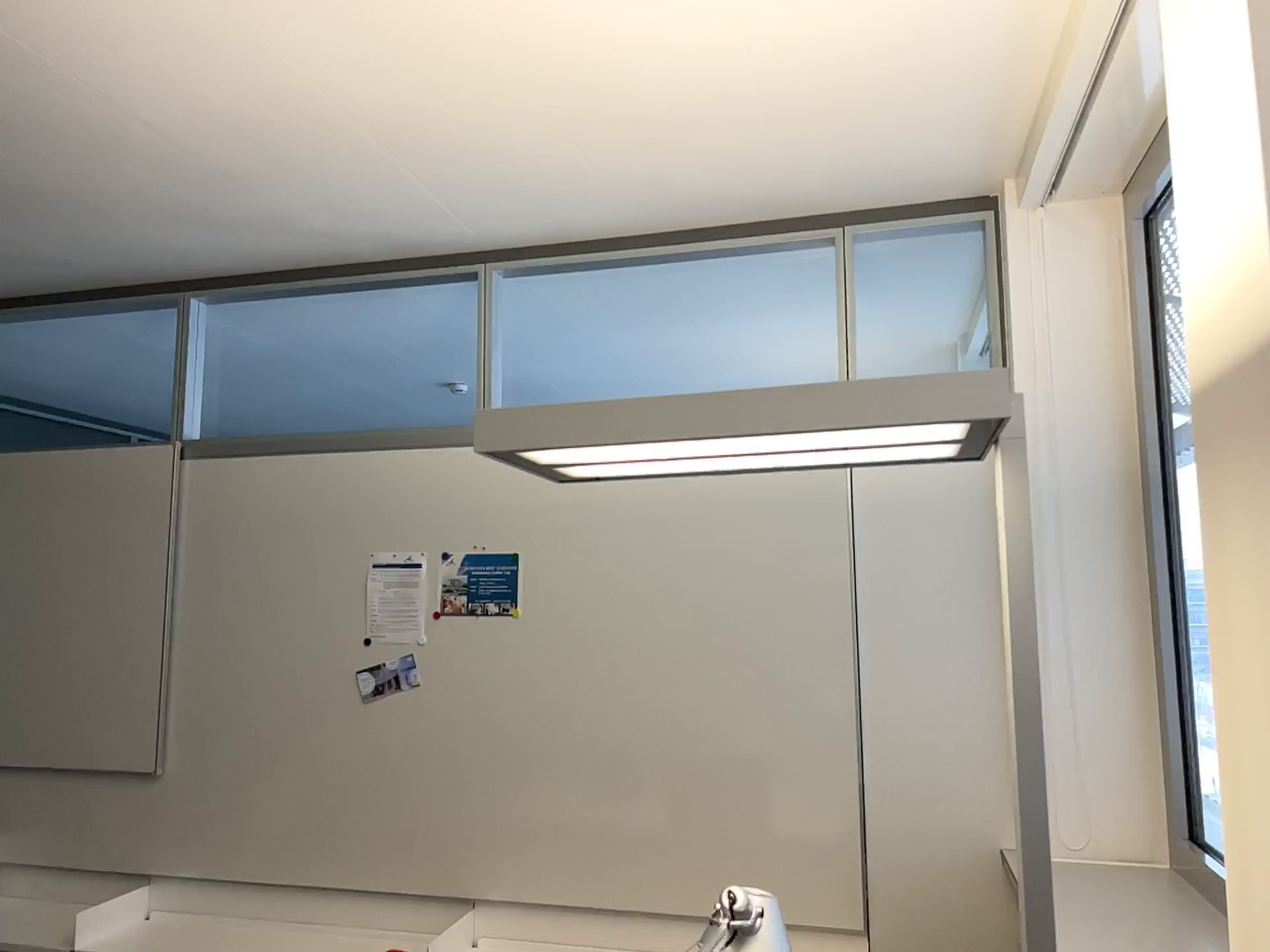}
        \\[-0.5mm]

        \rotatebox{90}{\tiny \method} &
        \includegraphics[width=.295\columnwidth]{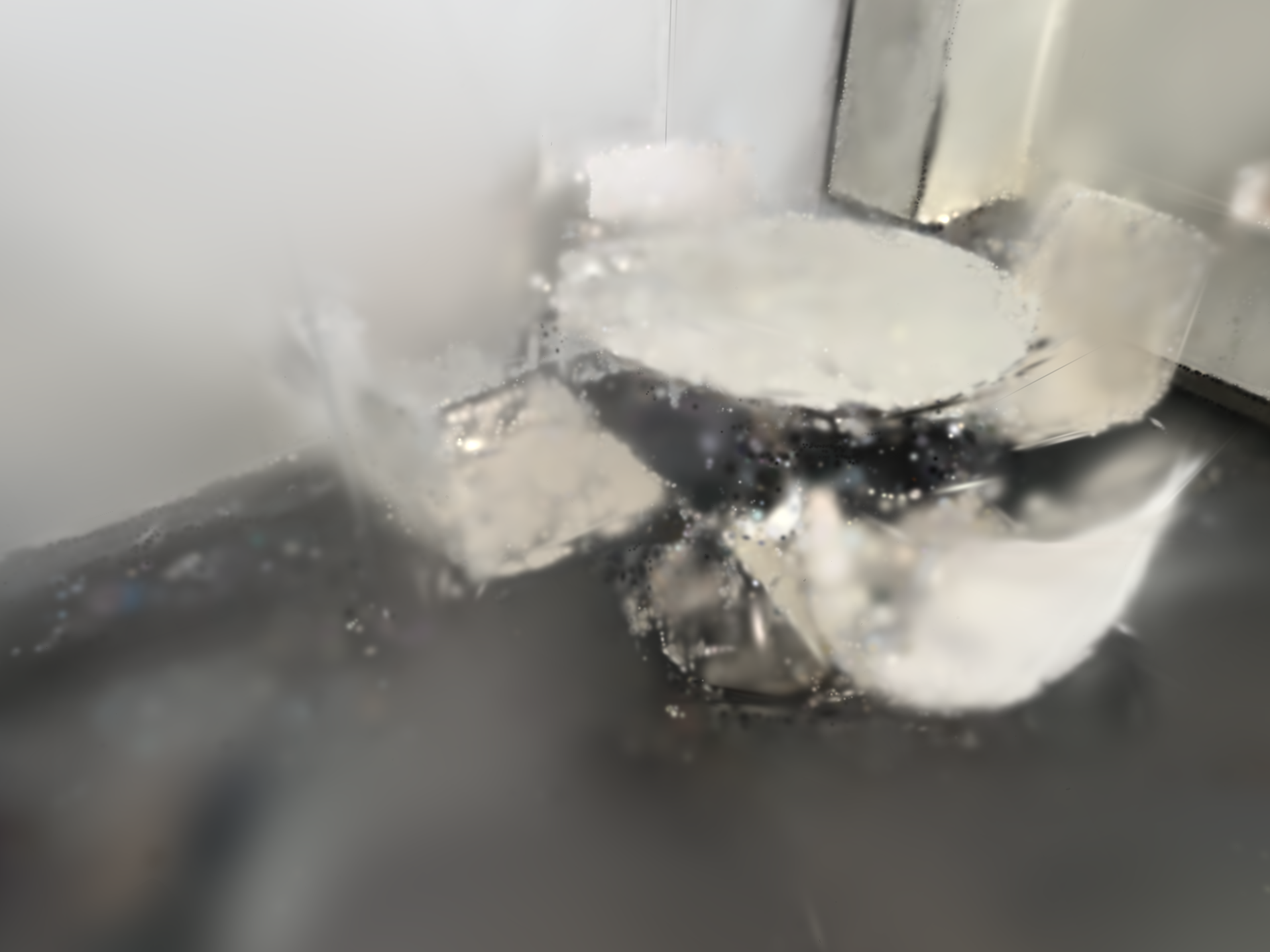} &
        \includegraphics[width=.295\columnwidth]{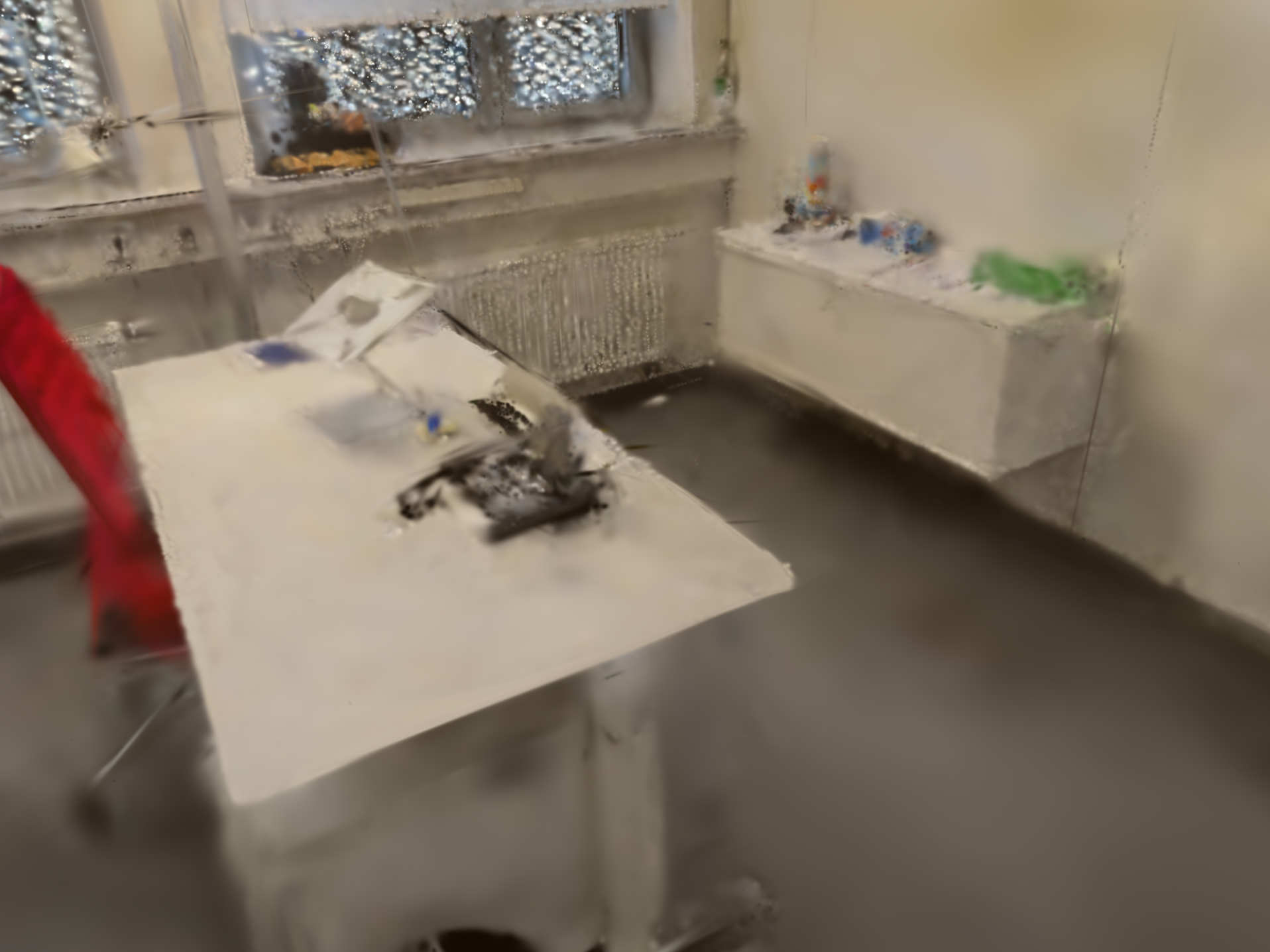} &
        \includegraphics[width=.295\columnwidth]{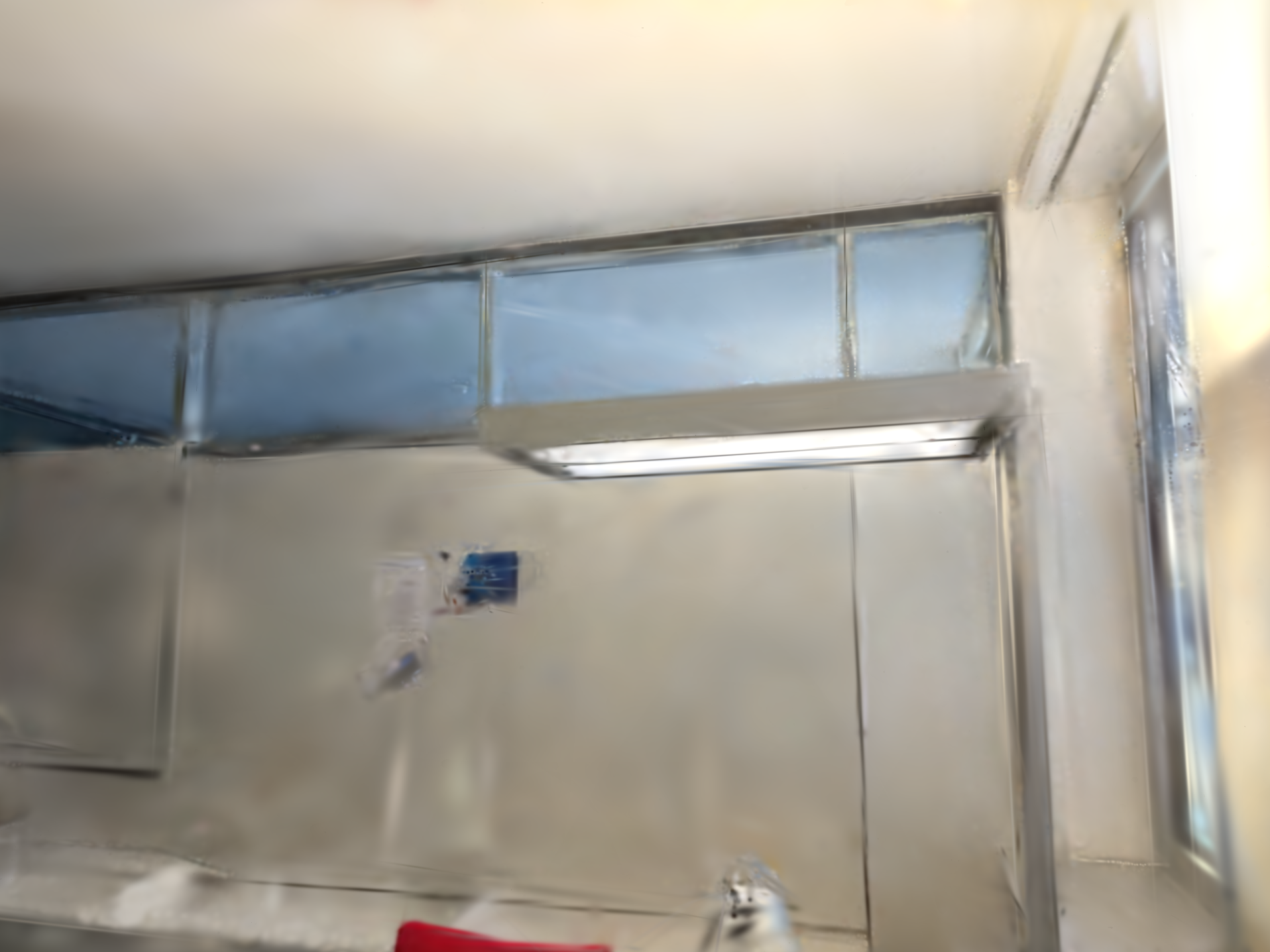}
    \end{tabular}

    \vspace{-1mm}
    \caption{
        \textbf{Qualitative reconstruction across additional indoor datasets.}
        Reference images and \method mapping-end renderings at representative
        viewpoints on \textbf{(a)} ScanNet and
        \textbf{(b)} ScanNet++ \texttt{8b5caf3398}.
    }
    \label{fig:cross_dataset_qualitative}
    \vspace{-2mm}
\end{figure}

\begin{figure}[t]
    \centering
    \setlength{\tabcolsep}{0pt}

    {\scriptsize\textbf{(a) TUM RGB-D
    \texttt{fr3/long\_office\_household}}}\par
    \vspace{0.7mm}

    {\scriptsize\textbf{RGB}}\par
    \vspace{0.3mm}

    \begin{tabular}{@{}*{3}{>{\centering\arraybackslash}m{0.333\linewidth}}@{}}
        \scriptsize GT &
        \scriptsize CaRtGS &
        \scriptsize \method
    \end{tabular}

    \vspace{-0.4mm}

    \includegraphics[
        width=\linewidth,
        trim=14.5pt 202pt 544.5pt 11.5pt,
        clip
    ]{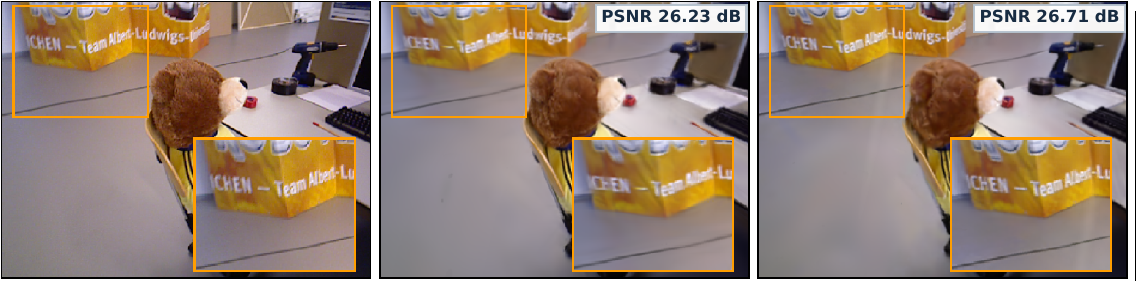}

    \vspace{1.2mm}

    {\scriptsize\textbf{Depth}}\par
    \vspace{0.3mm}

    \begin{tabular}{@{}*{3}{>{\centering\arraybackslash}m{0.333\linewidth}}@{}}
        \scriptsize GT &
        \scriptsize CaRtGS &
        \scriptsize \method
    \end{tabular}

    \vspace{-0.4mm}

    \includegraphics[
        width=\linewidth,
        trim=557.5pt 197.2pt 1.4pt 16.3pt,
        clip
    ]{figures/figure7_qualitative_rgb_depth.pdf}

    \vspace{1.8mm}

    {\scriptsize\textbf{(b) Real-sensor \texttt{kitchen1}}}\par
    \vspace{0.7mm}

    {\scriptsize\textbf{RGB}}\par
    \vspace{0.3mm}

    \begin{tabular}{@{}*{3}{>{\centering\arraybackslash}m{0.333\linewidth}}@{}}
        \scriptsize GT &
        \scriptsize CaRtGS &
        \scriptsize \method
    \end{tabular}

    \vspace{-0.4mm}

    \includegraphics[
        width=\linewidth,
        trim=14.5pt 36.4pt 544.5pt 176.5pt,
        clip
    ]{figures/figure7_qualitative_rgb_depth.pdf}

    \vspace{1.2mm}

    {\scriptsize\textbf{Depth}}\par
    \vspace{0.3mm}

    \begin{tabular}{@{}*{3}{>{\centering\arraybackslash}m{0.333\linewidth}}@{}}
        \scriptsize GT &
        \scriptsize CaRtGS &
        \scriptsize \method
    \end{tabular}

    \vspace{-0.4mm}

    \includegraphics[
        width=\linewidth,
        trim=557.5pt 15pt 1.4pt 181.8pt,
        clip
    ]{figures/figure7_qualitative_rgb_depth.pdf}

    \vspace{-1.6mm}
    {\tiny Depth [m]}\par

    \vspace{-0.5mm}

    \caption{
        \textbf{Qualitative RGB and depth reconstruction.}
        Reference, CaRtGS, and \method renderings from matched viewpoints
        on TUM RGB-D \texttt{fr3/long\_office\_household} and the
        real-sensor \texttt{kitchen1} sequence.
    }
    \label{fig:qualitative}
    \vspace{-2mm}
\end{figure}

Figures~\ref{fig:cross_dataset_qualitative} and~\ref{fig:qualitative}
provide complementary qualitative results. The former shows \method
mapping-end reconstructions on ScanNet and ScanNet++, while the latter compares
reference, CaRtGS, and \method RGB and depth renderings from matched viewpoints
on TUM RGB-D and the real-sensor \texttt{kitchen1} sequence.

\section{Discussion}
\label{sec:discussion}

\method treats continual RGB-D Gaussian mapping as a compute-allocation problem
across view selection, supervision fidelity, and geometry growth. The
experiments show that these decisions affect different parts of the
compute--quality trade-off: Load-Adaptive Fidelity reduces supervision work,
Targeted Geometry Growth controls where representation capacity is added, and Map-Guided View Scheduling redistributes optimization according to
regional reconstruction state. The same Gaussian map remains available for
further refinement after acquisition.
Real-time operation depends on the sustained workload and currently requires
manual tuning of several runtime parameters. Representation size also grows
with the explored scene. \method assumes a predominantly static environment
and a sufficiently consistent input trajectory; pose errors, dynamic objects,
and unreliable depth can still produce inconsistent geometry. Integrating a
visual odometry frontend would extend \method from mapping to a complete
RGB-D SLAM system.


\bibliographystyle{IEEEtran}
\bibliography{references}
\end{document}